\documentclass[runningheads]{llncs}
\usepackage{graphicx}
\usepackage{amsmath,amssymb} % define this before the line numbering.
\usepackage{color}
\usepackage[width=122mm,left=12mm,paperwidth=146mm,height=193mm,top=12mm,paperheight=217mm]{geometry}
\usepackage{booktabs}
\usepackage{multirow,rotating}
\usepackage{enumitem}
\usepackage{array}
\usepackage{tabularx}

\def\etal{\emph{et al.}}
\def\ie{\emph{i.e.}}
\def\eg{\emph{e.g.}}

\makeatletter
\def\blfootnote{\xdef\@thefnmark{}\@footnotetext}
\makeatother

\begin{document}
% \linenumbers
\pagestyle{headings}
\mainmatter

\title{Online Action Detection} % Replace with your title

\titlerunning{Online Action Detection}

\authorrunning{R. De Geest, E. Gavves, A. Ghodrati, Z. Li, C. Snoek, T. Tuytelaars}

\author{Roeland De Geest$^1$, Efstratios Gavves$^2$, Amir Ghodrati$^1$,\\ Zhenyang Li$^2$, Cees Snoek$^2$, Tinne Tuytelaars$^1$ }
\institute{$^1$KU Leuven - ESAT - PSI \hspace{1cm} $^2$University of Amsterdam - QUVA}
% First name: Roeland; Last name: De Geest
\institute{$^1$KU Leuven - ESAT - PSI\\
    \email{ \{roeland.degeest,amir.ghodrati,tinne.tuytelaars\}@esat.kuleuven.be} \\
    $^2$University of Amsterdam - QUVA\\
    \email{ \{e.gavves,z.li2,c.g.m.snoek\}@uva.nl}
}

\maketitle

\begin{abstract} 
In online action detection, the goal is to detect the start of an action in a video stream as soon as it happens. For instance, if a child is chasing a ball, an autonomous car should recognize what is going on and respond immediately. This is a very challenging problem for four reasons. First, only partial actions are observed. Second, there is a large variability in negative data. Third, the start of the action is unknown, so it is unclear over what time window the information should be integrated. Finally, in real world data, large within-class variability exists. This problem has been addressed before, but only to some extent.
Our contributions to online action detection are threefold. First, we introduce a realistic dataset composed of 27 episodes from 6 popular TV series. The dataset spans over 16 hours of footage annotated with 30 action classes, totaling 6,231 action instances. Second, we analyze and compare various baseline methods, showing this is a challenging problem for which none of the methods provides a good solution. Third, we analyze the change in performance when there is a variation in viewpoint, occlusion, truncation, etc.  We introduce an evaluation protocol for fair comparison. The dataset, the baselines and the models will all be made publicly available to encourage (much needed) further research on online action detection on realistic data.

\let\thefootnote\relax\footnote{This work was supported by the KU Leuven GOA project \emph{CAMETRON}.}
\addtocounter{footnote}{-1}\let\thefootnote\svthefootnote{}

\end{abstract}

\section{Introduction}
\begin{figure}[t]
\includegraphics[width=0.8\linewidth]{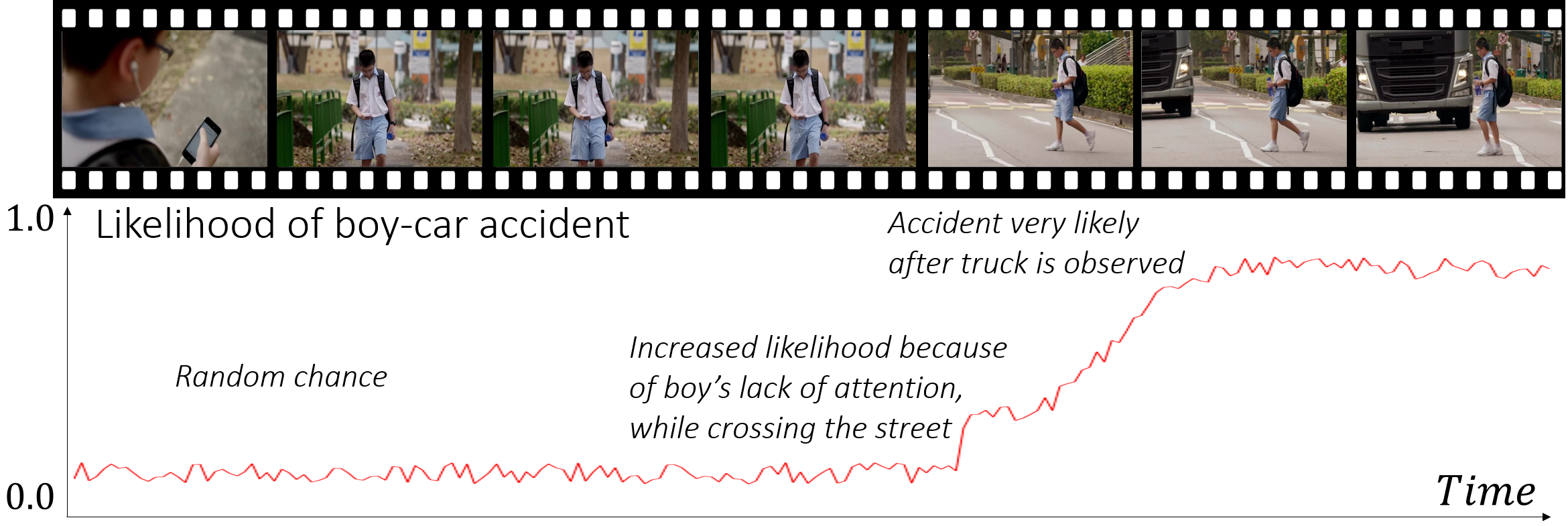}
\caption{Illustration of an online action detection prediction.
}
\label{fig:highfive}
\end{figure}

In this paper, we focus on the problem of {\em online action detection}.
Unlike traditional action recognition and action detection as studied in the literature to date, \eg,~\cite{wang2013action,Laptev2005,5995646,6909495,Fernando2016b,Bilen2016},
the goal of online action detection is to detect an action as it happens and ideally even before the action is fully completed.
Being able to detect an action at the time of the occurence can be useful in many practical applications
- think of a pro-active robot offering a helping hand; a surveillance camera raising an alarm
not just after the facts but well in time to allow for intervention;
a smart active camera system zooming in on the action scene and
recording it from the optimal perspective; or an autonomous car stopping for a child chasing a ball (see Figure~\ref{fig:highfive}).

A similar task coined `early event detection' has been brought to the attention of the community in the seminal work of Hoai and De la Torre~\cite{Hoai-DelaTorre-CVPR12,Hoai-DelaTorre-IJCV14}.
However, they consider only the special case of relatively short video fragments with the category label given as prior information.
Hence, it is assumed that it is known beforehand which action is going to take place. As the video is streamed, the system then only needs to
indicate, as early as possible but not too early, when the action has started. A further simplified setting, focusing more on classification instead of detection, has been studied in~\cite{6619187,6126349,Yu:2012:PHA:2393347.2396380,Kong,Tian}. In these works, the video starts with the onset of an action and ends when the action is completed. As the correct temporal segmentation is already provided, the system only needs to choose the most likely action out of a predefined set.

We claim these simplified setups are not representative for practical applications, where occurrences of any out of possibly many different action categories need to be detected in an online fashion, in (very) long video recordings with widely varying content. As we will show, this is a significantly more challenging task, to which the standard methods proposed in the literature provide only partial answers. Moreover, to date, no realistic benchmark dataset focusing on this problem has been released. In fact, the situation is somewhat reminiscent of the early days of action recognition, with datasets such as KTH~\cite{KTH} or Weizmann~\cite{Weizmann}. To alleviate this problem, we introduce the \textit{TVSeries} dataset, a new dataset consisting of 27 episodes of 6 popular TV series. The dataset is temporally annotated at the frame level w.r.t. 30 possible actions. Furthermore, metadata is added, containing extra information regarding the action occurrence, \eg,~whether the action instance is atypical compared to the rest of the action instances in the same class, occluded, or taken from an unusual viewpoint.

We mark several differences between \emph{online action detection} and `early event detection'. First, we think the term `event' should be preserved for longer term activities such as `baking a cake' or `changing a tire', as in the TrecVid MED challenge~\cite{over2014trecvid}, which, by the way, is more a retrieval task than a detection one. Second, for practical applications methods should process the video in an online fashion (as opposed to batch processing), preferably in realtime and with minimal latency. Hence we prefer the term `online' over `early'.

Given a streaming video as input, the system should output, ideally in realtime, whether the action is currently taking place (or not). This requires detecting the ongoing action as accurately as possible, no matter what is the stage of the action. Since we focus on longer videos, this task requires in turn discriminating the action from a variety of negative data, including both background frames as well as irrelevant actions. Realistic background frames do not depict prespecified `neutral' poses as in earlier datasets~\cite{Hoai-DelaTorre-IJCV14}. Similar to standard action detection, the wide variability and plethora of negative data makes the problem really challenging, although for online action detection the effects are even stronger. For a TV series episode with 20 minutes of footage, a typical `standing up' action might not be appearing for more than 10 seconds in total (less than 1\% of the total number of frames). Only if a method can cope with this data imbalance and the large variability in the negative data, it will be of any practical use. Additionally, given the streaming video as input, the method needs to decide the proper temporal window to pool information from for deriving the frame prediction. This is not trivial in an online setting, since the algorithm does not know starting and ending points bounding the action temporally.

In summary, the challenges of real-world online action detection are the following. First, actions need to be detected as soon as possible, ideally after only part of the action has been observed. Second, actions need to be detected from among a wide variety of irrelevant negative data. Third, starting from long, unsegmented video data, it is unclear what time window to pool information from.
Finally, we work with real world data, not artificially created for the purpose of action recognition. By design this results in large within-class variability.

Together with the TVSeries dataset, we propose an evaluation protocol, that allows comparing different solutions in a qualitative and quantitative manner. It is designed to be invariant to the number of instances of an action in the test set and less affected by the flux of negative data present in the videos.
Given this protocol we report initial results for a set of state-of-the-art baseline methods on this challenging task.
More specifically, we consider Fisher vectors~\cite{perronnin2010improving} with improved trajectories~\cite{wang2013action}, a deep ConvNet operating on a single frame basis \cite{SimonyanVeryDeep} and an LSTM network, recently popular for sequential modelling such as image captioning~\cite{JiaICCV2015} and action recognition~\cite{Donahue_2015_CVPR}, to encode the actions temporally.
As it turns out, detecting actions at the time of their occurrence in realistic settings, while keeping the number of false positives under control, is a much harder problem than one might conclude from results reported in the literature under more constrained settings, \eg,~offline action detection. With this new dataset and evaluation protocol, we hope to encourage more researchers to look into the challenging yet very practical task of \emph{online action detection}. 

In the next section, we discuss related work. In Section~\ref{sec:Dataset}, we describe the TVSeries dataset. Afterwards, we introduce our evaluation protocol. % in Section~\ref{Evaluation}. 
We evaluate several baselines and analyze their performance in Section~\ref{exp} and conclude in Section~\ref{conclusion}.

\section{Related work}

\noindent \textbf{Action detection datasets} 
The current datasets for action detection all have their limitations. In some datasets, \eg, UCF Sports~\cite{UCFSports}, the videos are temporally trimmed: they contain exactly the action, from start to finish. The task here is to find the spatial location of the action. However, in a video stream it is often more important to be able to localize an action in time, rather than in space. In surveillance, for instance, when a guard is alerted that something is happening, he looks at the screen and easily localizes the action.

Some action detection datasets only contain a limited amount of actions. MSRII~\cite{MSRII}, for example, contains only 54 short video sequences with only three action classes. The actions do not occur concurrently. The MPII Cooking Dataset~\cite{MPII} is larger: it has 44 videos with 65 actions. However, this dataset is recorded with a fixed camera and therefore every video contains only one shot and exactly the same background. Moreover, many actions are location dependent: \eg,~`Taking out of fridge' can only be done near the fridge. Occlusion is rare. Usually, the whole action is recorded and visible, from start to finish. 

Recently, some larger and more realistic datasets have been introduced. The Thumos detection challenge~\cite{THUMOS15} contains 24,000 (positive and negative) videos with 20 different actions; a similar dataset is FGA-240~\cite{SunMM15local}: it has 135,000 videos with 240 categories (85 sports, the rest fine-grained actions of these sports). In these datasets, all actions are sports related, so the background (the playing field) gives strong cues to help detection. The videos are downloaded from YouTube. As they are user created content, they often consist of only one shot: actions do not extend over multiple shots and are the main focus of the videos. Occlusions and partly recorded actions are rare. Another relevant dataset is ActivityNet~\cite{activitynet}. ActivityNet is larger and more varied and focuses on more generic categories, not just sports. The videos are downloaded from YouTube as well, so most have a duration between five and ten minutes. Since they are retrieved based on a textual query, it is very unlikely that one video contains multiple actions. Moreover, negative background data is likely class-specific as well. Therefore, action detection on this dataset is easier than the generic problem. Regarding datasets and online action detection, we experimentally make the observation that in realistic data, the negative background frames are by far the hardest obstacle for modeling the actions accurately. Hence, the aforementioned action datasets are not well suited for evaluating \emph{online action detection} reliably.\\

\noindent \textbf{Early action detection} Hoai and De la Torre~\cite{Hoai-DelaTorre-CVPR12,Hoai-DelaTorre-IJCV14} were the first to present `early event detection'. They simulate the sequential arrival of training data and train a structured output SVM, with the extra constraint that the output of frame $t+1$ should be higher than the output of frame $t$. At test time, they assume every video contains exactly one instance of a given action. As the video is streamed, the system starts detecting the action once a threshold is exceeded. Only at the end of the video, they decide on a specific start and end frame. In \cite{Hoai-DelaTorre-IJCV14} they discuss an extended setting where multiple actions per video are processed, however they never evaluate this. In our setting, we do not make any prior assumptions of the content of a video. Moreover, detecting the end of the action in an online fashion, as well as the start, is crucial.

\cite{Hoai-DelaTorre-CVPR12} uses three types of video data to test the method: sign language, facial expressions and simple actions from the Weizmann dataset~\cite{Weizmann}. The videos are all relatively short and look artificial: the person is centered and instructed to perform a specific action. In this work, we use realistic data and introduce a new dataset that is well-suited for online action detection. 
They also propose to use the ROC curve, AMOC curve and F1-score curve as evaluation metrics. As we will detail later, these metrics are not ideal for online action detection.

In a follow-up work, Huang \etal~\cite{HuangWYD14}, approach the problem more as classification than detection. They start assuming that every learned action can be happening, as well as a `non-action'. When more frames of the video are seen, the occurrence of some actions becomes more unlikely and they are discarded. When only one action remains, or no actions are removed for a certain amount of time, a detection happened. In their data, however, the non-action is very simple: a person is just standing. In the real-world data we use, the non-actions have very high variability and it is not easy to learn a model for them.
\\

\noindent\textbf{Offline action detection}
In this problem, the whole video is given. The task is to detect whether a given action occurs in this video, and if so, where it starts and ends (see \eg~\cite{5995646,6909495,Klaser,Tian:2013:SDP:2514950.2515975,wang2015exploring,gkioxari2015finding,yeung2015every}). Often the spatial location is determined as well. 
In this offline setting, the whole action can be observed first. Moreover, calculation time is not an issue. As a result, the best performing methods are often far too complicated to be used in a real-time setting.

A recent work by Yeung \etal~\cite{yeung2015end} explores action detection based on a limited number of frames. They train a recurrent neural network that takes a representation of a frame as input and selects another frame (at an arbitrary location in the video) to consider next. This way, they look at the most interesting frames only. In online detection, the goal is to detect an action based on a limited number of frames as well. However, the frames considered are always at the beginning of the action, while in~\cite{yeung2015end}, that is not necessarily the case: it is assumed the whole video is available and the RNN selects the interesting frames without constraints.
\\

\noindent \textbf{Early action classification} 
Another simplified setting, focusing on classification instead of detection,
has been studied in \eg~\cite{6619187,6126349,Yu:2012:PHA:2393347.2396380,Kong,Tian}. These works consider segmented actions. The system then only needs to choose one out of a predefined set of actions. A separate classifier is trained for every 10\%, 20\%, ..., 100\% of the video seen. During testing, it is known exactly how much percent of the action has been observed. This is clearly not valid in an online setting.

%\section{Online action detection}
\section{Dataset}
\label{sec:Dataset}

In this work, we introduce the TVSeries dataset. The videos in this dataset depict realistic actions as they happen in real life. Similar to the Hollywood2 dataset for action recognition~\cite{marszalek09}, our dataset is composed of professionally recorded videos. We annotated the first episodes of six recent TV series~\footnote{\emph{Breaking Bad} (3 episodes), \emph{How I Met Your Mother} (8), \emph{Mad Men} (3), \emph{Modern Family} (6), \emph{Sons of Anarchy} (3) and \emph{24} (4)}. We select the number of episodes such that we have around 150 minutes of every series: almost 16 hours in total. We divide the episodes over a training, validation and testing set. Every set contains at least one episode of every series: having different series in training and testing set would introduce a domain shift, and online action detection is already difficult enough by itself.

\begin{figure}[t]
\includegraphics[width=1.0\linewidth]{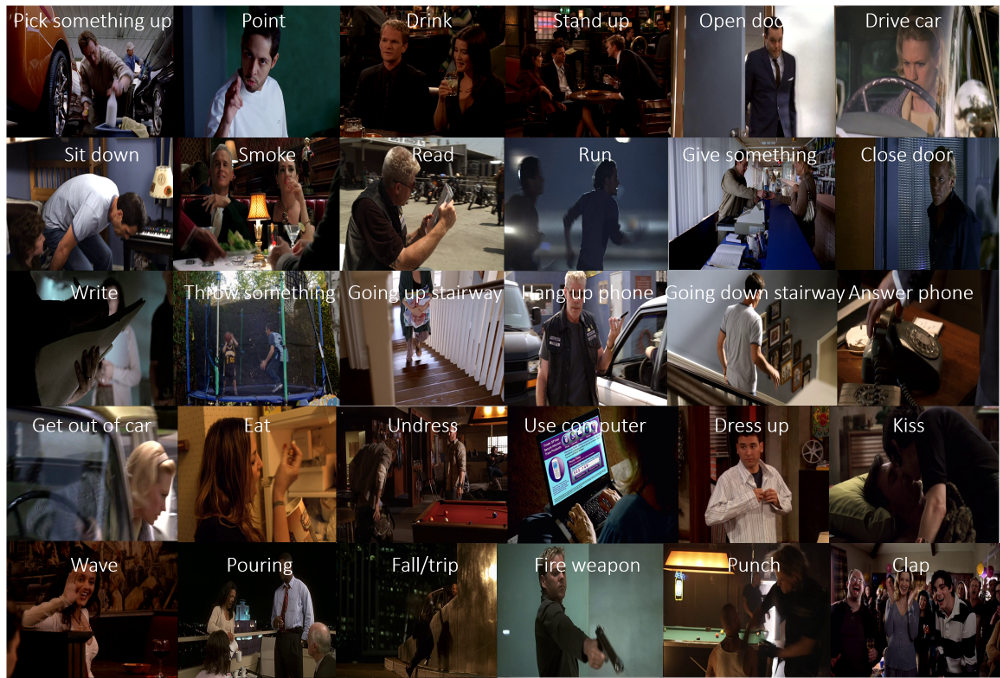}
\caption{A characteristic frame for each of the 30 classes in the \emph{TVSeries} dataset.}
\label{fig:dataset}
\end{figure}

We define 30 actions (see Table~\ref{tab:Online}). Every action occurs at least 50 times in the dataset. Annotations were done manually and afterwards checked by one person. The start of an action is defined as the first frame where one notices something is going to happen; the person is in rest position (or doing something completely different) in the previous frame. The end of an action is defined as the last frame that contains visual evidence of the action. After that, you can no longer tell that action has happened. The actions are only annotated temporally, not spatially.

\begin{table}[ht!]
\centering
\begin{tabular}{>{\raggedright}p{3.5cm} p{8cm} }
    \toprule
    \textbf{Dataset}& \\
    \emph{Source material} & 27 episodes of TV series: \emph{Breaking Bad}, \emph{How I Met Your Mother}, \emph{Mad Men}, \emph{Modern Family}, \emph{Sons of Anarchy}, \emph{24}. \\
\emph{Size} & ca. 16 hours \\
    \emph{Action classes number} & 30 \\
    \emph{Total number of actions} & 6,231 \\
    \midrule
    \textbf{Metadata}& \\
\emph{Atypical} & Does the actor perform the action in a way humans would call `atypical'? Example: `drinking' upside down. \\
\emph{Multiple persons} & Are multiple persons visible during the action? \\
\emph{Small or background} & Is the annotated action very small or in the background? \\
\emph{Side viewpoint} & Is (part of) the action recorded from the side? \\
\emph{Frontal viewpoint} & Is (part of) the action recorded from a frontal viewpoint? \\
\emph{Special viewpoint} & Is (part of) the action recorded from a special viewpoint? Example: `pouring' seen from the bottom of a glass. \\
\emph{Moving camera} & Is the camera moving during the action? \\
\emph{Shotcut} & Does the action instance extend over a shotcut? \\
\emph{Occlusion} & Is the part of the video where the action is (spatially) located occluded at some time during the action? \\
\emph{Spatial truncation} & Does part of the action extend beyond the frame borders? \\

\emph{Temporal truncation at the start} & Is the start of the action missing? \\
\emph{Temporal truncation at the end} & Is the end of the action missing? \\
    \midrule
    \multicolumn{2}{l}{\textbf{Automatically generated Metadata}} \\
    \emph{Length of action} & Actions divided in 4 quartiles based on number of frames \\
    \emph{Amount of motion} & Actions divided in 4 quartiles based on number of extracted improved trajectories \\
    \bottomrule
\end{tabular}
\caption{The TVSeries dataset and the specification of the provided metadata.}
\label{tab:dataset}
\end{table}

There is a large variability in this dataset. First, there are multiple actors, and everyone does an action his or her way. Second, different actions can occur at the same time, being performed by the same or multiple actors (as opposed to the easy setting of~\cite{Hoai-DelaTorre-IJCV14}, where actions are separated by a specific non-action). Third, the way the action is recorded can be very different. The viewpoint is not fixed. Part of the action can be occluded. In other cases, the recording only starts after the action has started, or it ends too early. Some of the actions are not crucial for the story in the series, and therefore, the director did not capture the actions clearly.  Other actions are performed by bystanders in the background and are very small. Fourth, the camera can be moving. Moreover, there are many shotcuts. Actions extend over multiple shots: the viewpoint of one action instance can suddenly change. Due to the long video sequences, containing multiple actions and a highly varying background, the shotcuts and the incomplete actions, this dataset is more challenging than the most realistic datasets currently used.

\begin{figure}[t]
\includegraphics[width=1.0\linewidth]{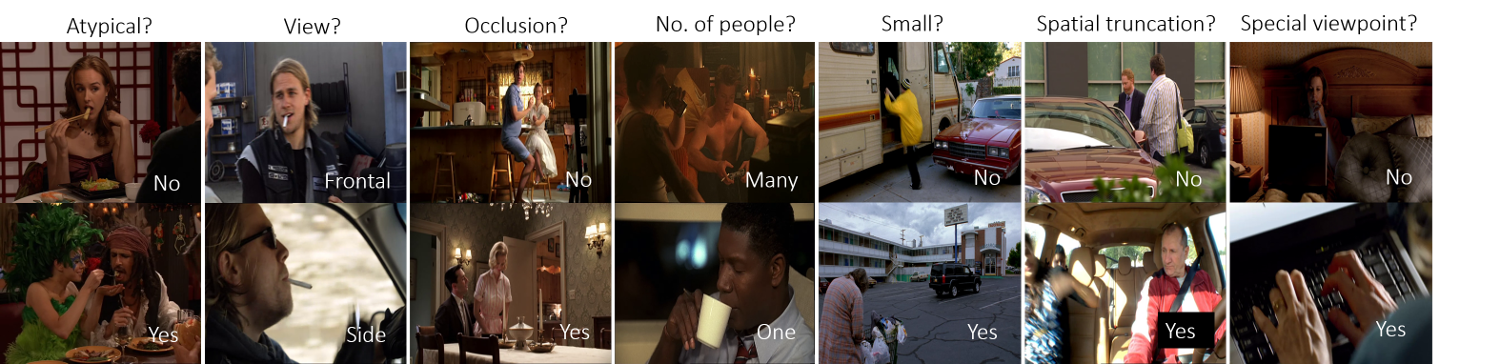}
\caption{
Example frames for some of the metadata annotations. Classes are `eat', `smoke', `stand up', `drink', `going up stairway', `get out of car' and `use computer'.
}
\label{fig:metadata}
\end{figure}

For every action instance, we provide metadata labels that give more information on how the action is performed and captured. In Table~\ref{tab:dataset} we summarize the dataset and the metadata, while in Fig.~\ref{fig:dataset} we present some characteristic frames from different classes.
In Fig.~\ref{fig:metadata} we show examples of metadata annotations.

The videos are ripped at a frame rate of 25 fps and have a resolution of 720 by 576 pixels. Some examples can be found in the supplemental material. This dataset will be made publicly available to encourage further research on (online) action detection on realistic data.

\section{Evaluation protocol}
\label{Evaluation}

\textbf{Relevant evaluation protocols} Existing evaluation protocols are not suited for the task of online action detection. In \emph{offline detection}, the main goal is to discover the start and end frame of an action, such that the detected action overlaps at least $\alpha$\% with the ground truth and the label of the detected action is correct~\cite{5995646,6909495,Klaser,Tian:2013:SDP:2514950.2515975}. A partial overlap cannot be distinguished from a full overlap, and it is unsure which part of the action is detected.
In \emph{early action classification}, temporally segmented actions are classified at points where 10\%, 20\%, ..., 100\% of the action is observed and the accuracies at these percents are measured~\cite{6619187,6126349,Yu:2012:PHA:2393347.2396380,Kong,Tian}. However, since it is a classification setting, this evaluation protocol cannot handle non-action intervals.

The evaluation metrics used for MMED~\cite{Hoai-DelaTorre-CVPR12,Hoai-DelaTorre-IJCV14} are the area under the ROC curve, the AMOC curve and the F1-score curve. The ROC curve shows, for different thresholds, the number of times a detector fires during the action (true positive rate, TPR) as a function of the number of times the detector fires before the action (false positive rate, FPR). The AMOC curve plots the average normalized time to detection (the percentage of the action that has been seen before the detector fires) as a function of the FPR for different thresholds. The F1-score curve tries to capture how well the method can localize the action. At every frame, the MMED method outputs the most probable start frame if an action ends at that frame. The F1-score is calculated at every \emph{action} frame, and this is plotted from 0-100\% of the action. 

These evaluation metrics are not really suited for online action detection. First, having three metrics instead of just one is sub-optimal. Second, every video gives rise to only one TP or FP. The assumption is made that a video contains the action exactly once. In a real-world streaming setting, this is obviously not the case. Finally, in an online action detection setting, methods do not need to label the start of the action in retrospect, after already having seen a sizable part of the action. The evaluation should therefore not consider a retrospective labeling of the action start, as the F1-score curve does.\\

\noindent \textbf{Proposed evaluation protocol} In online action detection, a decision needs to be made at every frame, for every action: how likely is it that the action is going on in that frame, based on the information available up to that point? Therefore, it is logical to use the average precision over all frames as a metric for the performance of an online action detector. 
First, the frames are ranked according to their confidence (high to low). The precision of a class at cut-off $k$ in this list is calculated as $Prec(k) = TP(k)/(TP(k)+FP(k))$
% \begin{equation} 
% \end{equation}
with $TP(k)$ the number of true positive frames and $FP(k)$ the number of false positives at the cut-off. The average precision of a class is then defined as $AP = \sum_{k}Prec(k)*I(k)/P$
% \begin{equation} 
% \end{equation}
with $I(k)$ an indicator function that is equal to 1 if frame $k$ is a true positive, and equal to 0 otherwise. $P$ is the total number of positive frames.
The mean of the AP over all classes (mAP) is then the final performance metric of an online action detection method.

This metric has one big disadvantage, though: it is sensitive to changes in the ratio of positive frames versus negative background frames (if the classifiers are not perfect), as discussed by Jeni \etal~\cite{jeni2013facing}. If there is (relatively speaking) more background data, the probability increases that some background frames are falsely detected with higher confidence than some true positives. So the AP will decrease. This makes it hard to compare the AP of two different classes when they do not have the same positive vs. negative ratio. Likewise, it makes it hard to evaluate performance on subsets of the data (\eg,~performance of unoccluded instances vs. occluded ones). To enable an easy, fair comparison, we introduce the {\em calibrated precision}:
\begin{equation} 
cPrec = \frac{TP}{TP+\frac{FP}{w}}=\frac{w*TP}{w*TP+FP}
\end{equation}
We choose $w$ equal to the ratio between negative frames and positive frames, such that the total weight of the negatives becomes equal to the total weight of the positives. Based on this calibrated precision, we can compute the {\em calibrated average precision (cAP)}, similar to the AP:
\begin{equation} 
cAP = \frac{\sum_{k}cPrec(k)*I(k)}{P}
\end{equation}
This way, the average precision is calculated as if there were an equal amount of positive and negative frames: the random score is 50\%. This evaluation metric is inspired by the work of Hoiem \etal~\cite{hoiem2012diagnosing}. They use a normalized average precision to compare object detection scores for different classes. Since in that case the number of negative data cannot be determined, they adjust the average precision as if every class has the same (arbitrary) amount of positive instances. Our calibrated average precision makes use of the number of negative data as well, and therefore, it is more suited for evaluation in our task.

For our dataset, we take the mAP as final performance measure. To compare the effectiveness of the different classifiers and the influence of the metadata labels, we use the cAP instead.

\section{Experiments}
\label{exp}

\subsection{Baseline features}
We analyze the difficulty of our dataset with three baseline methods. We opt for these, as they are the backbone of most action detection systems today.
\\

\noindent\textbf{1. Trajectories + FV} In our first approach, we use the improved trajectories of~\cite{wang2013action}, with default parameters. For every trajectory, we calculate the raw trajectory motion, and HOG, HOF and MBH around the trajectory. 
Based on these descriptors, we calculate Fisher vectors (FV)~\cite{perronnin2010improving} as in~\cite{wang2013action}.
These FVs are used as input for linear SVM classifiers: one one-vs-all SVM for every action class.
As examples for the SVM we use fixed-length windows, obtained as follows.
Our positive windows are the ones that are completely in a positive action instance, \ie, intersection of window and ground truth is equal to the length of the window. If the action is shorter than the window size, we take all windows that contain the action completely. As negative windows, we use windows of all other actions as well as background windows, where no action is happening. We train four SVMs for different window lengths: 20, 40, 60 and 80 frames. At test time, the prediction for the current frame is obtained by max-pooling the scores of windows of length 20, 40, 60 and 80 ending in the current frame.
\\

\noindent\textbf{2. CNN} As a second approach, we run a CNN on every frame separately. We choose the VGG-16 architecture \cite{SimonyanVeryDeep} which consists of 13 convolutional layers to train the RGB network, including a softmax layer to return class probabilities. Since our training data is relatively small, we first pre-train our model on UCF101 split-1, then we finetune on our dataset. We also do image flipping and multiscale cropping for data augmentation. As CNN relies on single frames only, there is no temporal information encoded.
\\

\noindent\textbf{3. LSTM} Our third approach is based on the recently successful LSTM~\cite{Donahue_2015_CVPR,yeung2015every}. LSTM is the most popular variant of recurrent neural networks, with a distinct ability of modeling better long and short term temporal patterns in sequence data, making them good candidates for modeling video data.
We use a single layer LSTM architecture with 512 hidden units. We directly resize each frame to 224x224 pixels (without data augmentation) and use it as input to extract the fc6 features from our CNN. These fc6 features are then fed into the LSTM. For training and testing, each video is split into multiple sequences of 16 frames (stride 1). 
Our LSTM model takes 16 frames as input at a time, and makes a prediction for the last frame.
The LSTM is connected with a softmax, which again returns class probabilities.

\subsection{Offline detection}

In offline detection, the goal is to find the start and end frame of any action that occurs in the video. All information of the video is available at once, and calculation time is not an issue.
As this is a more widely studied setting, we first report offline detection scores on our new dataset using the methods described above, as a reference.

To this end, the baselines need to be adapted to the offline setting. For baseline 1, we run the SVM classifiers over all windows of lengths 20, 40, 60 and 80. We then use a non-maximum suppression algorithm (as in~\cite{5995646}) to eliminate double detections. For baseline 2 and 3, we take a window around every frame and assign the score of that frame to the whole window. The length of the window is chosen for each class separately as the median of the duration of the instances of that class in the training set. We then use the same non-maximum suppression algorithm.

Evaluation is done in the traditional setting. Intersection over union is calculated between the detected windows and the ground truth. If this value is larger than an overlap ratio and the action class is correctly identified, the detection is considered correct. Then, the average precision is calculated. We obtain a mAP for overlap ratio 0.2 of 4.9\%, 1.1\% and 2.7\% for FV, CNN and LSTM respectively. The results for more overlap ratios and all classes separately can be found in the supplemental material.
% Old CNN and LSTM: 1.1\% and 2.4\%

In general, FVs are better than LSTM, which is better than CNN. The three methods perform best on different classes. FVs capture motion information, and are therefore best for classes that inherently have a lot of motion, like `stand up', `fall' and `punch', as opposed to actions like `write' and `eat'. CNN on the other hand is appearance-based, and therefore needs characteristic poses or context information from objects and scenes (`drive car', `read' and `drink' all provide these). The AP is lower than the AP of the FVs: with realistic data, this static information is not sufficient. LSTM uses the CNN features and is able to use their temporal order. This is not the same as having real motion information, but a step in the right direction (reflected by its score in between CNN and FV). It might be a good idea to use motion features (e.g. optical flow) as input for the LSTM, but testing that is beyond the scope of this paper.

The detection scores are quite low, indicating that this is a difficult dataset. For reference: the average classification accuracy of the actions (without taking the background into account), is 15.3\%, 24.7\% and 22.4\% for FV, CNN and LSTM. The FV score is lower than the other ones, likely because some action instances are so short that it is impossible to extract trajectories for them.

\begin{sidewaystable}[ph!]
\scalebox{0.6}{ %\scalebox{0.37}{
\centering
\begin{tabular}{l | c c c | c c c | c c c | c c c | c c c | c c c | c c c | c c c | c c c | c c c | c c c | c c c | c c c }  \hline
cAP (\%) & \multicolumn{3}{c}{Overall} & \multicolumn{3}{c}{\rotatebox{0}{A}} & \multicolumn{3}{c}{\rotatebox{0}{B}} & \multicolumn{3}{c}{\rotatebox{0}{C}} & \multicolumn{3}{c}{\rotatebox{0}{D}} & \multicolumn{3}{c}{\rotatebox{0}{E}} & \multicolumn{3}{c}{\rotatebox{0}{F}} & \multicolumn{3}{c}{\rotatebox{0}{G}} & \multicolumn{3}{c}{\rotatebox{0}{H}} & \multicolumn{3}{c}{\rotatebox{0}{I}} & \multicolumn{3}{c}{\rotatebox{0}{J}} & \multicolumn{3}{c}{\rotatebox{0}{K}} & \multicolumn{3}{c}{\rotatebox{0}{L}} \\ \hline
 & \rotatebox{90}{FV} & \rotatebox{90}{CNN} & \rotatebox{90}{LSTM} & \rotatebox{90}{FV} & \rotatebox{90}{CNN} & \rotatebox{90}{LSTM} & \rotatebox{90}{FV} & \rotatebox{90}{CNN} & \rotatebox{90}{LSTM} & \rotatebox{90}{FV} & \rotatebox{90}{CNN} & \rotatebox{90}{LSTM} & \rotatebox{90}{FV} & \rotatebox{90}{CNN} & \rotatebox{90}{LSTM} & \rotatebox{90}{FV} & \rotatebox{90}{CNN} & \rotatebox{90}{LSTM} & \rotatebox{90}{FV} & \rotatebox{90}{CNN} & \rotatebox{90}{LSTM} & \rotatebox{90}{FV} & \rotatebox{90}{CNN} & \rotatebox{90}{LSTM} & \rotatebox{90}{FV} & \rotatebox{90}{CNN} & \rotatebox{90}{LSTM} & \rotatebox{90}{FV} & \rotatebox{90}{CNN} & \rotatebox{90}{LSTM} & \rotatebox{90}{FV} & \rotatebox{90}{CNN} & \rotatebox{90}{LSTM} & \rotatebox{90}{FV} & \rotatebox{90}{CNN} & \rotatebox{90}{LSTM} & \rotatebox{90}{FV} & \rotatebox{90}{CNN} & \rotatebox{90}{LSTM} \\ \hline

\textit{Pick s/th up}
 & 70.0 & 56.1 & 68.1 & -- & -- & -- & -1.8 & -3.2 & -12.3 & -22.4 & -3.4 & -26.5 & 5.1 & -5.9 & -10.3 & -3.7 & 3.9 & 9.2 & -6.1 & 5.3 & 1.6 & -2.8 & -8.1 & 8.3 & -2.9 & -2.2 & 8.8 & -1.7 & -0.8 & -4.3 & -8.9 & -13.5 & 1.2 & -11.9 & -12.8 & -3.1 & -10.7 & -2.0 & -2.3 \\
\textit{Point}
 & 67.4 & 53.9 & 53.1 & -- & -- & -- & 7.3 & -12.3 & -3.0 & -- & -- & -- & -15.9 & -16.9 & -8.1 & 17.8 & 16.9 & 8.7 & -- & -- & -- & 1.7 & -13.8 & -1.3 & 3.2 & -0.4 & -9.1 & 0.5 & -9.0 & -3.2 & -5.5 & 2.6 & -0.9 & 2.4 & -11.9 & 1.2 & 11.0 & 1.5 & 6.7 \\
\textit{Drink}
 & 87.7 & 73.3 & 79.7 & -- & -- & -- & 3.6 & 15.1 & 17.3 & -7.6 & 9.4 & -1.9 & 3.1 & 6.5 & 1.5 & -0.9 & -5.8 & -0.5 & -- & -- & -- & -8.8 & -24.0 & -11.6 & 9.7 & 13.1 & 9.1 & -6.2 & -6.2 & -3.1 & -4.1 & 4.8 & 3.6 & -3.2 & 1.6 & -1.2 & -5.5 & 3.1 & -0.0 \\
\textit{Stand up}
 & 81.3 & 52.9 & 58.8 & -2.6 & 2.2 & -12.0 & -5.0 & 0.9 & -0.0 & -9.8 & -11.4 & -21.2 & -0.8 & -13.2 & -7.9 & 1.8 & 7.7 & 11.4 & -- & -- & -- & 5.9 & 1.5 & 18.5 & -1.8 & -1.2 & -2.5 & -7.0 & -10.8 & -13.7 & 4.6 & 7.2 & 10.8 & 1.4 & 3.6 & 2.6 & -8.4 & 11.5 & -15.8 \\
\textit{Run}
 & 88.2 & 58.6 & 72.9 & -- & -- & -- & 11.5 & 1.8 & -2.2 & 0.9 & 8.0 & 10.7 & -3.1 & -6.4 & -8.3 & 0.7 & 14.8 & 6.6 & 2.3 & -8.2 & 9.4 & 4.4 & -1.2 & -5.7 & -1.2 & 18.1 & -6.9 & 1.2 & -1.8 & 7.9 & 1.7 & -4.9 & -11.7 & -2.0 & -14.2 & -13.5 & -0.5 & -8.8 & -7.0 \\
\textit{Sit down}
 & 77.4 & 45.6 & 49.1 & -12.0 & 3.8 & -6.9 & -0.4 & -1.4 & 2.0 & -20.3 & 1.0 & -12.0 & -1.9 & -1.9 & -4.8 & 10.4 & 4.5 & 12.3 & -- & -- & -- & 12.1 & 2.0 & 4.2 & 8.4 & 0.3 & 8.2 & -8.3 & 1.2 & -0.2 & 6.1 & -3.1 & 7.3 & -5.1 & 2.1 & 1.4 & -8.0 & -1.7 & 8.7 \\
\textit{Read}
 & 64.6 & 60.3 & 61.8 & -- & -- & -- & 5.2 & 16.7 & 9.6 & 7.2 & 13.2 & -0.2 & -8.5 & -18.0 & -13.9 & -3.1 & -10.1 & -12.3 & -- & -- & -- & 1.2 & 4.1 & 5.9 & -2.5 & -19.4 & -18.6 & 1.9 & 3.5 & -5.7 & -7.0 & -13.2 & -15.6 & 1.0 & -2.3 & -5.2 & 5.1 & 10.9 & 3.4 \\
\textit{Smoke}
 & 79.2 & 66.7 & 67.1 & -0.7 & -10.4 & 7.6 & 18.4 & 1.7 & 6.2 & 11.5 & -4.0 & -2.5 & 3.5 & -9.8 & -9.6 & -3.4 & 14.2 & 12.2 & -- & -- & -- & 1.9 & 7.6 & 6.6 & -- & -- & -- & 10.5 & -7.5 & -13.7 & 10.5 & -12.1 & -1.1 & -6.5 & -5.4 & -9.9 & 5.0 & -8.6 & -9.7 \\
\textit{Drive car}
 & 91.5 & 80.7 & 90.3 & -- & -- & -- & 18.4 & -15.3 & 1.8 & -- & -- & -- & -15.5 & 5.9 & -3.2 & 4.1 & -3.4 & 7.7 & -- & -- & -- & -13.1 & 34.1 & 2.3 & -13.7 & 22.6 & 4.2 & -- & -- & -- & 0.4 & 14.1 & 10.0 & 7.0 & 0.7 & -4.1 & -0.0 & -8.0 & 2.8 \\
\textit{Open door}
 & 80.9 & 51.6 & 73.1 & 0.4 & 24.5 & 3.6 & -9.6 & -6.7 & -14.3 & -24.5 & 12.3 & -11.1 & -1.3 & -4.5 & -8.4 & 0.1 & 9.2 & 6.0 & -- & -- & -- & 15.9 & 7.4 & 1.8 & -0.2 & 1.1 & -5.1 & -5.0 & -6.7 & -7.9 & 12.3 & 2.1 & 5.6 & -12.1 & -7.1 & -0.6 & 12.2 & 8.0 & 7.2 \\
\textit{Give s/th}
 & 74.9 & 53.5 & 52.2 & -- & -- & -- & -- & -- & -- & -- & -- & -- & -6.5 & -5.0 & 5.4 & 8.1 & 6.2 & -1.0 & -- & -- & -- & -12.6 & 4.0 & -4.8 & -6.5 & 6.4 & 8.9 & -14.1 & 13.8 & -7.5 & 2.6 & 5.8 & 6.7 & -0.9 & -10.7 & -4.8 & -- & -- & -- \\
\textit{Use computer}
 & 70.2 & 77.8 & 83.9 & -- & -- & -- & -4.1 & 5.6 & 7.5 & -- & -- & -- & -0.6 & 3.9 & -7.1 & 9.5 & 4.8 & -1.8 & -- & -- & -- & -- & -- & -- & -12.4 & 1.9 & 12.9 & 4.9 & -0.2 & 4.8 & -7.7 & -11.6 & -6.8 & 15.7 & -5.3 & -9.9 & 6.7 & -3.3 & -10.5 \\
\textit{Write}
 & 47.2 & 46.2 & 62.3 & -- & -- & -- & 7.8 & 4.8 & 2.2 & -5.5 & 33.6 & -14.8 & 1.4 & -16.0 & -21.1 & -3.0 & 5.3 & 9.4 & -4.1 & 7.1 & 10.8 & -7.4 & 19.8 & 2.2 & -6.2 & -26.6 & -28.3 & 6.2 & 9.4 & -5.0 & -10.2 & -6.9 & -10.6 & -8.6 & -21.7 & -7.2 & -1.1 & 12.3 & 0.5 \\
\textit{Stairway down}
 & 83.7 & 65.2 & 68.2 & -- & -- & -- & -15.5 & -15.3 & -17.7 & -21.1 & -13.8 & -18.7 & 6.6 & 16.4 & 12.9 & 20.0 & 6.3 & 5.0 & -- & -- & -- & -- & -- & -- & -- & -- & -- & 3.1 & 9.8 & 4.0 & -6.7 & -12.0 & 3.4 & -16.0 & -9.9 & -6.1 & -16.2 & -14.0 & -15.9 \\
\textit{Close door}
 & 81.7 & 54.2 & 73.7 & -- & -- & -- & -4.5 & 1.1 & -0.7 & -0.5 & 8.3 & -4.8 & 2.7 & -1.8 & -0.7 & 1.1 & 10.3 & 4.7 & -- & -- & -- & 10.3 & 7.0 & -7.6 & -7.2 & 10.4 & -0.9 & 2.6 & 0.7 & 16.5 & -5.9 & 0.9 & -18.8 & -2.1 & 11.8 & -5.0 & -3.8 & 1.5 & -18.0 \\
\textit{Stairway up}
 & 66.1 & 53.6 & 69.8 & -- & -- & -- & -- & -- & -- & -2.6 & -1.6 & -13.8 & -11.0 & 0.4 & 2.8 & 15.7 & -0.7 & 10.7 & -- & -- & -- & -1.7 & 3.3 & 22.6 & -- & -- & -- & 12.6 & -5.1 & 0.1 & -16.4 & 13.3 & -11.1 & -11.1 & -2.1 & -4.9 & -10.6 & -0.3 & 1.6 \\
\textit{Throw s/th}
 & 68.5 & 53.1 & 55.2 & -- & -- & -- & 18.0 & -6.3 & -1.1 & -- & -- & -- & 6.3 & -14.4 & -9.0 & -3.1 & 12.3 & 9.7 & -- & -- & -- & -8.8 & 18.7 & -3.3 & -- & -- & -- & 5.5 & -0.1 & -1.7 & 9.3 & 8.7 & -9.8 & 2.2 & 26.0 & -4.3 & -- & -- & -- \\
\textit{Get in/out car}
 & 89.7 & 84.7 & 79.0 & -- & -- & -- & -10.1 & -13.7 & -8.7 & -- & -- & -- & 3.3 & 5.9 & -8.1 & -2.2 & -5.7 & 5.7 & -- & -- & -- & -- & -- & -- & -- & -- & -- & -- & -- & -- & 5.1 & 8.7 & 9.6 & -11.7 & -9.6 & 4.0 & -- & -- & -- \\
\textit{Hang up phone}
 & 61.5 & 52.4 & 47.1 & -- & -- & -- & -16.5 & 13.6 & 3.6 & -- & -- & -- & -30.7 & -6.4 & -10.4 & 30.7 & 6.4 & 10.4 & -- & -- & -- & -- & -- & -- & -- & -- & -- & -- & -- & -- & -17.3 & -18.8 & 4.0 & -- & -- & -- & -- & -- & -- \\
\textit{Eat}
 & 63.7 & 65.0 & 52.2 & -- & -- & -- & 21.1 & 12.0 & 7.5 & 14.6 & 11.3 & 13.0 & -2.8 & 1.5 & 12.2 & 0.2 & -10.1 & 2.8 & -- & -- & -- & -8.2 & -27.7 & 3.1 & -- & -- & -- & 1.0 & -3.5 & 8.9 & 0.1 & 5.8 & 0.9 & -2.0 & 4.5 & -7.2 & -4.1 & 11.8 & -2.3 \\
\textit{Answer phone}
 & 57.3 & 46.4 & 46.2 & -- & -- & -- & 11.4 & -29.0 & -2.3 & -- & -- & -- & -6.5 & 5.3 & 6.9 & 15.9 & -25.7 & -19.5 & -- & -- & -- & -- & -- & -- & -- & -- & -- & -- & -- & -- & -- & -- & -- & -- & -- & -- & -- & -- & -- \\
\textit{Clap}
 & 69.7 & 59.0 & 79.6 & -- & -- & -- & -- & -- & -- & 1.7 & -13.8 & -5.8 & -0.4 & -15.2 & -6.2 & -- & -- & -- & -- & -- & -- & 13.0 & 10.1 & 34.2 & -- & -- & -- & 7.6 & -18.8 & 0.8 & 15.0 & -0.3 & 15.4 & 4.0 & 2.3 & 15.4 & 8.0 & 3.3 & 5.3 \\
\textit{Dress up}
 & 65.0 & 45.1 & 56.3 & -- & -- & -- & 6.2 & -16.6 & 12.8 & -- & -- & -- & -24.7 & 8.5 & -4.5 & 1.2 & -9.8 & 0.8 & -- & -- & -- & -- & -- & -- & -17.9 & 7.5 & -3.8 & -1.4 & 4.7 & -2.8 & 6.6 & 15.3 & 6.4 & -11.0 & 0.1 & -17.7 & -8.9 & 3.2 & -5.4 \\
\textit{Undress}
 & 67.8 & 67.4 & 55.4 & -- & -- & -- & -- & -- & -- & -9.2 & -3.3 & -8.0 & 6.0 & -3.9 & 8.3 & -1.1 & 8.9 & -4.5 & -- & -- & -- & -- & -- & -- & -- & -- & -- & 7.1 & 9.3 & -4.2 & 12.7 & -16.5 & -3.5 & -- & -- & -- & -1.9 & 3.5 & 8.6 \\
\textit{Kiss}
 & 71.4 & 51.2 & 66.3 & -- & -- & -- & -- & -- & -- & -- & -- & -- & -- & -- & -- & -- & -- & -- & -- & -- & -- & -- & -- & -- & -- & -- & -- & -8.1 & 1.0 & -7.3 & -- & -- & -- & -- & -- & -- & -- & -- & -- \\
\textit{Fall/trip}
 & 89.6 & 50.4 & 70.3 & -5.9 & 3.7 & 12.0 & -- & -- & -- & -- & -- & -- & 9.2 & 2.2 & -12.6 & -5.1 & -3.8 & 14.6 & -- & -- & -- & -- & -- & -- & -8.8 & -16.2 & -9.0 & 2.7 & 4.2 & -19.2 & -0.4 & -4.1 & 15.1 & -- & -- & -- & 7.0 & 19.7 & 17.5 \\
\textit{Wave}
 & 57.9 & 62.3 & 55.3 & -- & -- & -- & -- & -- & -- & -- & -- & -- & 6.0 & 4.8 & 5.9 & -17.6 & -13.2 & 1.0 & -- & -- & -- & -17.9 & -14.3 & 22.3 & -- & -- & -- & -- & -- & -- & -- & -- & -- & -- & -- & -- & 0.0 & -16.9 & 15.7 \\
\textit{Pour}
 & 74.4 & 76.2 & 68.5 & -6.4 & 37.7 & -15.8 & -1.6 & -20.4 & -23.1 & -- & -- & -- & 12.8 & -31.4 & 12.1 & -11.9 & 17.7 & -9.1 & -- & -- & -- & -- & -- & -- & -- & -- & -- & -- & -- & -- & -6.4 & 37.7 & -15.8 & -6.1 & 18.2 & 16.7 & -2.0 & -7.5 & 14.1 \\
\textit{Punch}
 & 93.7 & 68.5 & 46.5 & -- & -- & -- & -- & -- & -- & -- & -- & -- & 0.6 & -0.0 & 10.0 & -5.2 & 3.2 & -7.3 & -- & -- & -- & -- & -- & -- & -- & -- & -- & 0.5 & -0.3 & -9.4 & -1.2 & 0.3 & -6.4 & 5.2 & 0.1 & 3.6 & -4.3 & -2.0 & 4.9 \\
\textit{Fire weapon}
 & 87.4 & 90.9 & 61.8 & -- & -- & -- & -- & -- & -- & -- & -- & -- & -- & -- & -- & -- & -- & -- & -- & -- & -- & -- & -- & -- & -- & -- & -- & -- & -- & -- & -- & -- & -- & -- & -- & -- & -- & -- & -- \\
\hline
\textit{\textbf{Mean}}
 & \textbf{74.3} & \textbf{60.8} & \textbf{64.1} & \textbf{-4.5} & \textbf{10.2} & \textbf{-1.9} & \textbf{2.8} & \textbf{-3.2} & \textbf{-0.7} & \textbf{-5.8} & \textbf{3.0} & \textbf{-7.8} & \textbf{-2.3} & \textbf{-3.9} & \textbf{-2.7} & \textbf{2.9} & \textbf{2.4} & \textbf{3.4} & \textbf{-2.7} & \textbf{1.4} & \textbf{7.3} & \textbf{-0.8} & \textbf{1.7} & \textbf{5.4} & \textbf{-4.0} & \textbf{1.0} & \textbf{-2.1} & \textbf{0.7} & \textbf{-0.6} & \textbf{-2.9} & \textbf{-0.4} & \textbf{0.4} & \textbf{-0.5} & \textbf{-3.1} & \textbf{-1.8} & \textbf{-2.6} & \textbf{-1.3} & \textbf{0.7} & \textbf{0.4} \\ \hline

\end{tabular}}
\caption{Calibrated average precision for online action detection. Actions are sorted according to number of instances in the dataset from high to low. For the metadata labels: the difference between the calibrated precision of `yes' and `no' is shown. We mark with $(-)$ the entries with less than 5 `yes and 5 `no' action instances, as results would be unreliable. A: Atypical, B: Multiple persons, C: Small or background, D: Side viewpoint, E: Frontal viewpoint, F: Special viewpoint, G: Moving camera, H: Shotcut, I: Occlusion, J: Spatial truncation, K: Temporal truncation at the start, L: Temporal truncation at the end.}
\label{tab:Online}
\end{sidewaystable}

\subsection{Online detection}

In online detection, we decide at every moment whether a specific action is happening \emph{now}. This decision can not use information of the next frames, since this information is not yet available.
We evaluate by reporting the average precision over frames, as discussed in Section~\ref{Evaluation}. The mAP is 5.2\%, 1.9\% and 2.7\% for the FV, CNN and LSTM respectively. The values are very low, because the amount of negative data is very high, but still clearly better than the random mAP of 0.7\%. Here too, FVs score higher than LSTM and CNN. However, FVs computed on dense trajectories are slower. Dense trajectories, which occupy most of the computations, have a computational complexity of about $\mathcal{O}(S D^2 k f^2 +\mathcal{V})$, for $S$ scales and $D$ average frame width and height, employing $k$ convolutional kernels of size $f$ for smoothing and spatio-temporal gradients used in HOG/HOF/MBH and, $\mathcal{V}$ the computational complexity of the respective optical flow algorithm used. In practice using FVs from the features computed on dense trajectories is hard in a realtime setting. In comparison, CNN have a complexity of $\mathcal{O}(\sum_i^L C_i M_i^2 f_i^2)$ assuming an $L$-layered network with $C_i$ channels, $M_i$ feature map size (on average considerably smaller than $D$) and $f_i$ filter size and a thresholding (ReLU) non-linearity, while $\mathcal{O}\Big (\sum_i^L C_i M_i^2 f_i^2+\sum_t \phi (M_t u) \Big )$ for LSTMs that receive CNN feature maps as input, considering $u$ memory units and $t$ timesteps and non-linearities with complexity $\mathcal{\phi}$. Most importantly, because of the recursive nature of matrix multiplications, neural network based models are largely parallelizable in GPU architectures, allowing for much faster computations. 
%They require about 9 million multiply-accumulate operations per frame, substantially more than CNN (\roeland{TODO}) and LSTM (\roeland{TODO}).
% Old CNN and LSTM: 1.7\% and 2.1\%

To be able to compare the scores of the different classes, we calculate the cAP (see Table~\ref{tab:Online}). Multiple classifiers perform close to the random value of 50\, especially the CNN.
The conclusions for offline detection are valid here as well. FVs are best for actions that intrinsically have a lot of motion (`run', `punch'), while CNN needs context information and characteristic poses for its best classes (`fire weapon', `get in/out car').

Table~\ref{tab:overtime} shows the mean cAP for frames in every ten-percent interval of actions. FVs need some time to collect information of trajectories in windows. Their performance reaches its maximum near the end of the action. The cAP of the other methods is constant for all frames.

\begin{table}[t]
\scalebox{0.8}{
\centering
\begin{tabular}{l c c c c c c c c c c }  \hline
mean cAP (\%) 
 & 0-10\% & 10-20\% & 20-30\% & 30-40\% & 40-50\% & 50-60\% & 60-70\% & 70-80\% & 80-90\% & 90-100\% \\ \hline
\textit{FV}
 & 67.0 & 68.4 & 69.9 & 71.3 & 73.0 & 74.0 & 75.0 & 76.4 & 76.5 & 76.8 \\
 \textit{CNN}
% & 60.0 & 60.4 & 60.4 & 60.5 & 60.6 & 60.5 & 60.8 & 60.8 & 61.1 & 61.0  \\
 & 61.0 & 61.0 & 61.2 & 61.1 & 61.2 & 61.2 & 61.3 & 61.5 & 61.4 & 61.5  \\
 \textit{LSTM}
 & 63.3 & 64.5 & 64.5 & 64.3 & 65.0 & 64.7 & 64.4 & 64.3 & 64.4 & 64.3 \\ \hline
 \end{tabular}}
 \caption{Mean cAP for different baselines when only a part of every action is considered: first $10\%$ frames of the action, next $10\%$ \ldots last $10\%$, vs. all frames not containing the considered action.}
\label{tab:overtime}
\end{table}

\subsection{Metadata analysis}

We do an analysis based on the different metadata provided with the dataset. To be able to derive some meaningful conclusions, we just select those action categories for which 
we have at least 5 action instances in each of the two splits (\eg,~classes that have at least 5 atypical and 5 typical instances). The results are presented in Table \ref{tab:Online}. The most interesting observations are discussed below.

\noindent\textbf{Multiple persons} When there are multiple persons in the scene, the performance of FV slightly improves. The highest increase occurs with actions like `throw something' and `eat', which generally are performed in group. In contrast, actions like `hang up phone' and `close door' are recognized less often. For CNN, the average performance decreases when there are more persons present. When one person is present in the image (instead of a group of people), action-specific context is stronger. This explains the reduced performance.

%Column C
\noindent\textbf{Small or background} The FVs are clearly not capable of capturing the motion of small persons. The trajectories are hard to extract. Moreover, there are few of them, so their contribution to the FVs is relatively limited. CNN relies more on the context that is present in the whole image and is less sensitive to changes in size. In fact, when the action is small, more context may be available.

% Column D and E
\noindent\textbf{Side and frontal viewpoint} Analyzing the mean does not make sense here: the definitions of `frontal' and `side' depend on the action class. Interesting to note is that the performances of the three classifiers change differently for different actions. When one of them increases, there often is another one that decreases. The classifiers capture different information, and therefore, combining them seems a good idea to obtain better results.

% Column H
\noindent\textbf{Shotcut} Both temporal methods are negatively affected by shotcuts. Trajectories for FVs are interrupted and discarded, and it takes 15 frames to generate new ones. Therefore, some frames have less information. LSTM combines information from multiple frames. If there is a shotcut, the relation between the frames is not as clear. On the other hand, CNN uses only the current frame, so its accuracy does not change much.
%\noindent\textbf{Shotcut} All methods are negatively affected by shotcuts. Trajectories for FVs are interrupted and discarded, and it takes 15 frames to generate new ones. Therefore, some frames have less information. CNN does not use information of multiple frames, yet its accuracy decreases. A possible explanation for this could come from a cinematographic point of view. If an action is important for the story, it is shown relatively clear and large, and often not interrupted by shotcuts that could confuse the audience. Therefore, the relative number of clear actions is higher when there is no shotcut. An appearance-based method, like CNN, benefits from this.

% Column J - TODO
%\noindent\textbf{Spatial truncation} In this case, a part of the action is not visible. The accuracy of all methods decreases. CNN has most problems. This too can be explained from a cinematographic point of view. In TV series, often only the upper body is captured. For actions where a person is moving, like `going down stairway', this implies the absence of characteristic leg poses. FVs still capture the downward motion of the upper body `going down the stairway', or the shaking of the shoulders when a person is `running'.

% Column K and L - TODO
\noindent\textbf{Temporal truncation at start and end} For the temporal methods, the performance is worse when the start of the action is missing. These methods use information from previous frames. If an action is shorter because the beginning is missing, it takes relatively speaking more time before they have constructed a good representation. It does not matter that much whether the end of the action is missing.

\section{Conclusion}
\label{conclusion}
Online action detection is a difficult problem, that has not been studied in a real-world setting and with realistic data before. There are four main challenges. First, only partial actions are available (as previously stressed in~\cite{Hoai-DelaTorre-CVPR12,Hoai-DelaTorre-IJCV14,HuangWYD14}). Second, the negative data is highly variable and should not give rise to many false positives. Third, the start frame of an action is not known beforehand, so it is unclear over what time window to integrate the information. Fourth, large within-class variability exists in real-world data.

We collected a new dataset 
%, with metadata that indicates interesting characteristics of the actions, 
and proposed an evaluation protocol to assist the research on online action detection. We tested a few baselines and showed none of the simple methods perform well. A realistic setting is clearly different from the artificial setups that were previously used in an online action detection context. Therefore, online action detection is a novel problem far from being solved, as existing methodologies fall short on delivering reliable results.

\bibliographystyle{splncs}
\bibliography{oad}
\end{document}